\documentclass[journal,twocolumn]{IEEEtran}
\usepackage{graphicx,times,amsmath,amssymb,subfigure,stfloats,threeparttable,booktabs,bm,multirow,cite,csquotes}
\usepackage[lined,ruled,commentsnumbered]{algorithm2e}
\allowdisplaybreaks

\makeatletter

\newcommand{\Rmnum}[1]{\expandafter\@slowromancap\romannumeral #1@}

\makeatother

\begin{document}

\title{MBGD-RDA Training and Rule Pruning for Concise TSK Fuzzy Regression Models}

\author{Dongrui~Wu\\
Key Laboratory of the Ministry of Education for Image Processing and Intelligent Control, School of Artificial Intelligence and Automation, Huazhong University of Science and Technology, Wuhan 430074, China.\\
Email: drwu@hust.edu.cn.}

\maketitle

\begin{abstract}
To effectively train Takagi-Sugeno-Kang (TSK) fuzzy systems for regression problems, a Mini-Batch Gradient Descent with Regularization, DropRule, and AdaBound (MBGD-RDA) algorithm was recently proposed. It has demonstrated superior performances; however, there are also some limitations, e.g., it does not allow the user to specify the number of rules directly, and only Gaussian MFs can be used. This paper proposes two variants of MBGD-RDA to remedy these limitations, and show that they outperform the original MBGD-RDA and the classical ANFIS algorithms with the same number of rules. Furthermore, we also propose a rule pruning algorithm for TSK fuzzy systems, which can reduce the number of rules without significantly sacrificing the regression performance. Experiments showed that the rules obtained from pruning are generally better than training them from scratch directly, especially when Gaussian MFs are used.
\end{abstract}


\IEEEpeerreviewmaketitle

\section{Introduction}

Takagi-Sugeno-Kang (TSK) fuzzy systems \cite{Nguyen2019} have been used successfully in numerous applications. Its reasoning is based on IF-THEN rules, which is easier to interpret, compared with other black-box machine learning models such as neural networks. However, training a TSK fuzzy system is not easy, especially when the dataset is large. Traditional training approaches, e.g., evolutionary algorithms \cite{drwuEAAI2006}, batch gradient descent \cite{Wang1992b,Rumelhart1986}, and gradient descent plus least squares estimation (LSE) \cite{Jang1993}, all suffer from various problems \cite{drwuGD2020}.

Inspired by the connections between TSK fuzzy systems and neural networks~\cite{drwuTSK2020}, a Mini-Batch Gradient Descent with Regularization, DropRule, and AdaBound (MBGD-RDA) algorithm for training TSK fuzzy regression models has recently been proposed \cite{drwuGD2020}. It borrows many concepts from deep learning~\cite{Goodfellow2016}, e.g., MBGD to handle big data, regularization and DropRule (inspired by DropOut~\cite{Srivastava2014} and DropConnect~\cite{Wan2013}) to improve generalization, and AdaBound~\cite{Luo2019} to speed-up the training. It may be the only available TSK regression model training algorithm that can effectively deal with big data\footnote{Recently, an MBGD with uniform regularization and batch normalization algorithm \cite{drwuBN2020} was also proposed to deal with big data classification problems.}.

However, MBGD-RDA still has several limitations:
\begin{enumerate}
\item \emph{Computational cost}: MBGD-RDA needs to specify the number of Gaussian membership functions (MFs) in each input domain. Assume there are $M$ inputs, and the $m$-th input domain has $M_m$ MFs. Then, the total number of rules is $\prod_{m=1}^M M_m$, which may be prohibitively large when $M_m$ and/or $M$ is large. \cite{drwuGD2020} deals with this problem by using principal component analysis (PCA) \cite{Jolliffe2002} to reduce the number of feature dimensionality from $M$ to at most 5. However, information may be lost during this process, and the final regression precision is hence affected.

\item \emph{Interpretability}: Interpretability is a major advantage of fuzzy systems over other black-box machine learning models. However, as the number of rules increases, the interpretability rapidly decreases. PCA may be used to reduce the feature dimensionality, and hence the number of rules. However, the principal component features are different from the original features, which increases the difficulty in understanding.

\item \emph{Flexibility}: $\prod_{m=1}^M M_m$, the total number of rules, can assume a very limited number of feasible values. For example, when $M=5$, the smallest number of rules is 32, achieved when $M_m=2$ for all $m$. The next smallest number of rules is 48, achieved when one input has $M_m=3$ and all others have $M_m=2$. In practice the user may want to specify the number of rules as an arbitrary value, e.g., 10, 20, etc. This is not achievable using the current approach.

\item \emph{Types of MFs}: \cite{drwuGD2020} only considers Gaussian MFs, whereas sometimes people may prefer trapezoidal MFs.
\end{enumerate}

This paper proposes two variants of MBGD-RDA and a rule pruning algorithm for them. It makes the following contributions:
\begin{enumerate}
\item To reduce the computational cost and increase the interpretability and flexibility of MBGD-RDA, we extend MBGD-RDA to a more flexible form, which allows the user to specify the number of rules directly. It can use both Gaussian and trapezoidal MFs.
\item We propose a simple yet effective rule pruning approach for TSK fuzzy systems, based on the MBGD-RDA variants. This solves an important problem in practice: the user may not know \emph{a priori} how many rules should be used to achieve a good compromise between regression performance and rulebase simplicity.   So, he/she can specify a relatively large number of rules at the beginning, and then use our rule pruning algorithm to automatically prune the rulebase.
\item Experiments show that our rule pruning approach can not only reduce the number of rules, but also often achieve better performance than training from a reduced number of rules directly. For example, starting from 30 rules, our algorithm may tell that only 15 of them are necessary, and outputs a TSK fuzzy regression model with 15 rules. This 15-rule TSK fuzzy system, obtained from rule pruning, often achieves better regression performance than training a TSK fuzzy system with 15 rules directly.
\end{enumerate}

The remainder of this paper is organized as follows: Section~\ref{sect:variants} proposes the two variants of MBGD-RDA. Section~\ref{sect:pruning} describes the rule pruning algorithm. Section~\ref{sect:experiments} presents experiment results to validate the performance of the rule pruning algorithm. Finally, Section~\ref{sect:conclusions} draws conclusions.

\section{Variants of MBGD-RDA} \label{sect:variants}

This section introduces two variants of MBGD-RDA, which allow the user to specify the number of rules directly, instead of the number of MFs in each input domain. The first variant uses Gaussian MFs, and the second uses trapezoidal MFs. The key notations are summarized in Table~\ref{tab:annotations}, which are mostly identical to those in \cite{drwuGD2020}.

\begin{table}[htbp] \centering  \setlength{\tabcolsep}{2mm}
\caption{Key notations used in this paper.}   \label{tab:annotations}
\begin{tabular}{c|l}   \toprule
Notation    &\multicolumn{1}{|c}{Definition} \\ \midrule
$N$ & Number of labeled training samples \\
$M$ & Number of features \\
$R$ & Number of rules  \\
$\mathbf{x}_n=(x_{n,1},$ & \multirow{2}{*}{The $n$th training sample} \\
$\ ...,x_{n,M})^T$ &  \\
$y_n$ & Groundtruth output corresponding to $\mathbf{x}_n$ \\
$X_{r,m}$ & MF for the $m$th feature in the $r$th rule \\
$w_{r,0},...,w_{r,M}$ & Consequent parameters of the $r$th rule\\
$y_r(\mathbf{x}_n)$ & Output of the $r$th rule for $\mathbf{x}_n$\\
$\mu_{X_{r,m}}(x_{n,m})$ & Membership grade of $x_{n,m}$ on $X_{r,m}$\\
$f_r(\mathbf{x}_n)$ & Firing level of $\mathbf{x}_n$ on the $r$th rule \\
$y(\mathbf{x}_n)$ & Output of the TSK fuzzy system for $\mathbf{x}_n$ \\
$L$ & $\ell_2$ regularized loss function \\
$\lambda$   & $\ell_2$ regularization coefficient  \\
$M_m$          & Number of MFs in each input domain\\
$N_{bs}$        & Mini-batch size \\
$K$       & Number of training epochs  \\
$\alpha$        & Initial learning rate \\
$P$       & DropRule rate \\ \bottomrule
\end{tabular}
\end{table}

First, we briefly introduce the original MBGD-RDA algorithm proposed in \cite{drwuGD2020}.

\subsection{The TSK Fuzzy Regression Model}

Assume the input $\mathbf{x}=(x_1,...,x_M)^T\in \mathbb{R}^{M\times1}$, and the TSK fuzzy system has $R$ rules:
\begin{align}
\mathrm{Rule}_r: &\mbox{ IF } x_1 \mbox{ is } X_{r,1} \mbox{ and } \cdots \mbox{ and } x_M \mbox{ is } X_{r,M},\nonumber\\
& \mbox{ THEN } y_r(\mathbf{x})=w_{r,0}+\sum_{m=1}^M w_{r,m}x_m, \label{eq:Rr}
\end{align}
where $X_{r,m}$ ($r=1,...,R$; $m=1,...,M$) are fuzzy sets, and $w_{r,0}$ and $w_{r,m}$ are consequent parameters.

Let $\mu_{X_{r,m}}(x_{m})$ be the membership grade of $x_m$ on $X_{r,m}$. The firing level of $\mathrm{Rule}_r$ is:
\begin{align}
f_r(\mathbf{x})=\prod_{m=1}^M\mu_{X_{r,m}}(x_{m}), \label{eq:fr}
\end{align}
and the output of the TSK fuzzy system is:
\begin{align}
y(\mathbf{x})=\frac{\sum_{r=1}^R f_r(\mathbf{x})y_r(\mathbf{x})}{\sum_{r=1}^R f_r(\mathbf{x})}. \label{eq:yTSK}
\end{align}
Or, if we define the normalized firing levels as:
\begin{align}
\bar{f}_r(\mathbf{x})=\frac{f_r(\mathbf{x})}{\sum_{k=1}^Rf_k(\mathbf{x})},\quad r=1,...,R \label{eq:f}
\end{align}
then, (\ref{eq:yTSK}) can be rewritten as:
\begin{align}
y(\mathbf{x})&=\sum_{r=1}^R\bar{f}_r(\mathbf{x})\cdot y_r(\mathbf{x}). \label{eq:yTSK2}
\end{align}

\subsection{Mini-Batch Gradient Descent (MBGD)}

Assume there are $N$ training samples $\{\mathbf{x}_n,y_n\}_{n=1}^N$, where $\mathbf{x}_n=(x_{n,1},...,x_{n,M})^T\in \mathbb{R}^{M\times1}$. MBGD randomly samples $N_{bs}\in[1, N]$ training samples, computes the gradients from them, and then updates the antecedent and consequent parameters of the TSK fuzzy system.

Let $\bm{\theta}_k$ be the model parameter vector in the $k$th training epoch, and $\partial{L}/\partial{\bm{\theta}_k}$ the first gradients of the loss function $L$. Then, the update rule is:
\begin{align}
\bm{\theta}_k=\bm{\theta}_{k-1}-\alpha\frac{\partial{L}}{\partial{\bm{\theta}_{k-1}}},
\end{align}
where $\alpha>0$ is the learning rate (step size).

\subsection{Regularization}

MBGD-RDA uses the following $\ell_2$ regularized loss function:
\begin{align}
L=\frac{1}{2}\sum_{n=1}^{N_{bs}}\left[y_n-y(\mathbf{x}_n)\right]^2+\frac{\lambda }{2}\sum_{r=1}^R\sum_{m=1}^M w_{r,m}^2, \label{eq:loss}
\end{align}
where $\lambda\ge 0$ is a regularization parameter. Note that $w_{r,0}$ ($r=1,...,R$) are not regularized in (\ref{eq:loss}).

\subsection{DropRule}

DropOut \cite{Srivastava2014} is a common technique for reducing overfitting and improving generalization in deep learning. It randomly discards some neurons and their connections during the training.

Khalifa and Frigui \cite{Khalifa2016} were the first to introduce the DropOut concept to the training of fuzzy classifiers. They called it \emph{Rule Dropout}. Let the DropOut rate be $P\in(0,1)$. In training, they first compute the normalized firing levels of all rules, discard each rule with probability $(1-P)$, and then use gradient descent to update the parameters of the remaining rules. In test, all rules are used, but the output is scaled by $P$.

A new DropRule approach \cite{drwuGD2020} with reduced computational cost and simpler operation was recently introduced for TSK fuzzy regression models. For each training sample, one sets the firing level of a rule to its true firing level with probability $P$, and to zero with probability $1-P$, equivalent to dropping that rule. Then, MBGD is used to update the parameters of the rules that are not dropped. When the training is done, all rules are used in computing the output for a new input, just as in a traditional TSK fuzzy system. Because the rules are dropped before computing the normalized firing levels, no scaling is needed in test.

\subsection{AdaBound}

Adam \cite{Kingma2015}, used almost everywhere in deep learning, adjusts the individualized learning rate for each parameter adaptively. This may result in better training and generalization performance than using a fixed learning rate. AdaBound \cite{Luo2019} improves Adam by bounding the learning rates so that they cannot be too large nor too small. At the beginning of the training, the bound is $[0,+\infty)$. As the training goes on, the bound approaches $[0.01,0.01]$.

\subsection{MBGD-RDA Using Gaussian MFs}

The membership grade of $x_m$ on a Gaussian MF $X_{r,m}$ is:
\begin{align}
\mu_{X_{r,m}}(x_{m})=\exp\left(-\frac{(x_{m}-c_{r,m})^2}{2\sigma_{r,m}^2}\right), \label{eq:mu}
\end{align}
where $c_{r,m}$ is the center of the Gaussian MF, and $\sigma_{r,m}$ the standard deviation.

When Gaussian MFs are used, the gradients of the loss function (\ref{eq:loss}) are given in (\ref{eq:dEodC})-(\ref{eq:dEdB}), where $x_{n,0}\equiv1$, and $I(m)$ is an indicator function:
\begin{figure*}[htpb]
\begin{align}
\frac{\partial L}{\partial c_{r,m}}&=\frac{1}{2}\sum_{n=1}^{N_{bs}}\frac{\partial L}{\partial y(\mathbf{x}_n)}
\frac{\partial y(\mathbf{x}_n)}{\partial f_r(\mathbf{x}_n)}
\frac{\partial f_r(\mathbf{x}_n)}{\partial \mu_{X_{r,m}}(x_{n,m})}
\frac{\partial \mu_{X_{r,m}}(x_{n,m})}{\partial c_{r,m}}\nonumber\\
&=\sum_{n=1}^{N_{bs}}\left[\left(y(\mathbf{x}_n)-y_n\right)
\frac{y_r(\mathbf{x}_n)\sum_{i=1}^Rf_i(\mathbf{x}_n)
-\sum_{i=1}^Rf_i(\mathbf{x}_n)y_i(\mathbf{x}_n)}
{\left[\sum_{i=1}^Rf_i(\mathbf{x}_n)\right]^2}f_r(\mathbf{x}_n)
\frac{x_{n,m}-c_{r,m}}{\sigma_{r,m}^2}\right] \label{eq:dEodC}\\
\frac{\partial L}{\partial \sigma_{r,m}}&=\frac{1}{2}\sum_{n=1}^{N_{bs}}\frac{\partial L}{\partial y(\mathbf{x}_n)}
\frac{\partial y(\mathbf{x}_n)}{\partial f_r(\mathbf{x}_n)}
\frac{\partial f_r(\mathbf{x}_n)}{\partial \mu_{X_{r,m}}(x_{n,m})}
\frac{\partial \mu_{X_{r,m}}(x_{n,m})}{\partial \sigma_{r,m}}\nonumber\\
&=\sum_{n=1}^{N_{bs}}\left[\left(y(\mathbf{x}_n)-y_n\right)
\frac{y_r(\mathbf{x}_n)\sum_{i=1}^Rf_i(\mathbf{x}_n)
-\sum_{i=1}^Rf_i(\mathbf{x}_n)y_i(\mathbf{x}_n)}
{\left[\sum_{i=1}^Rf_i(\mathbf{x}_n)\right]^2}f_r(\mathbf{x}_n)
\frac{(x_{n,m}-c_{r,m})^2}{\sigma_{r,m}^3}\right] \label{eq:dEodSigma}\\
\frac{\partial L}{\partial w_{r,m}}&=\frac{1}{2}\sum_{n=1}^{N_{bs}}\frac{\partial L}{\partial y(\mathbf{x}_n)}
\frac{\partial y(\mathbf{x}_n)}{\partial y_r(\mathbf{x}_n)}
\frac{\partial y_r(\mathbf{x}_n)}{\partial w_{r,m}}+\frac{\lambda}{2} \frac{\partial L}{\partial w_{r,m}}=\sum_{n=1}^{N_{bs}}\left[\left(y(\mathbf{x}_n)-y_n\right) \frac{f_r(\mathbf{x}_n)}{\sum_{i=1}^Rf_i(\mathbf{x}_n)}\cdot x_{n,m}\right] +\lambda I(m)w_{r,m}\label{eq:dEdB}
\end{align}
\end{figure*}
\begin{align}
I(m)=\left\{\begin{array}{cc}
              0, & m=0 \\
              1, & m>0
            \end{array}\right..
\end{align}
$I(m)$ ensures that $w_{r,0}$ ($r=1,...,R$) are not regularized.

The pseudo-code of the MBGD-RDA variant using Gaussian MFs is shown in Algorithm~\ref{alg:MBGD-RDA}. Compared with the original MBGD-RDA algorithm in \cite{drwuGD2020}, it has two main changes: 1) here we specify the total number of TSK rules, instead of the number of MFs in each input domain; and, b) fuzzy $c$-means clustering \cite{Bezdek1981} initialization\footnote{We also tested $k$-means clustering initialization; however, it performed much worse than fuzzy $c$-means clustering initialization.} of the rules, instead of a semi-random initialization, is used.

\begin{algorithm*}[htbp] 
\KwIn{$N$ labeled training samples $\{\mathbf{x}_n,y_n\}_{n=1}^N$, where $\mathbf{x}_n=(x_{n,1},...,x_{n,M})^T\in \mathbb{R}^{M\times1}$; $L(\bm{\theta})$, the loss function for the TSK fuzzy system parameter vector $\bm{\theta}$; $R$, the number of rules; $K$, the maximum number of training epochs; $N_{bs}\in[1,N]$, the mini-batch size; $P\in(0,1)$, the DropRule rate; $\alpha$, the initial learning rate (step size); $\lambda$, the $\ell_2$ regularization coefficient;  Optional: $\bm{\theta}_0$, the initial $\bm{\theta}$.}
\KwOut{The final $\bm{\theta}$.}
\If{$\bm{\theta}_0$ is not supplied}{
\tcp{Fuzzy $c$-means clustering initialization}
Perform fuzzy $c$-means clustering ($c=R$) on $\{\bm{x}_n\}_{n=1}^N$\;
Denote the $r$-th cluster center as $\bar{\bm{c}}_r=[\bar{c}_{r,1},...,\bar{c}_{r,M}]$, and the corresponding fuzzy partition as $\bm{u}_r=[u_{r,1},...,u_{r,N}]$, $r=1,...,R$\;
\For{$r=1,...,R$}{
Initialize $w_{r,0}=\sum_{n=1}^N y_n u_{r,n}\left/\sum_{n=1}^Nu_{r,n}\right.$\;
\For{$m=1,...,M$}{
Initialize $w_{r,m}=0$, $c_{r,m}=\bar{c}_{r,m}$, and $\sigma_{r,m}$ as $\bm{u}_r$ weighted standard deviation of $\{x_{n,m}\}_{n=1}^N$\;}}
$\bm{\theta}_0$ is the concatenation of all $c_{r,m}$, $\sigma_{r,m}$, $w_{r,0}$ and $w_{r,m}$\;}
\tcp{Update $\bm{\theta}$}
$\mathbf{m}_0=\bm{0}$;
$\mathbf{v}_0=\bm{0}$\;
\For{$k=1,...,K$}
{
Randomly select $N_{bs}$ training samples\;
\For{$n=1,...,N_{bs}$}{
\For{$r=1,...,R$}{
\tcp{DropRule}
$f_r(\mathbf{x}_n)=0$\;
Generate $p$, a uniformly distributed random number in $[0,1]$\;
\If{$p\le P$}{
Compute $f_r(\mathbf{x}_n)$, the firing level of $\mathbf{x}_n$ on $\mathrm{Rule}_r$\;}}
Compute $y(\mathbf{x}_n)$, the TSK fuzzy system output for $\mathbf{x}_n$, by (\ref{eq:yTSK})\;
\For{each element $\bm{\theta}_{k-1}(i)$ in $\bm{\theta}_{k-1}$}{
\begin{align*}
\mathbf{g}_k(i)=\left\{\begin{array}{ll}
                                       \frac{\partial L}{\partial \bm{\theta}_{k-1}(i)}, & \mbox{if }\bm{\theta}_{k-1}(i) \mbox{ was used in computing } y(\mathbf{x}_n) \\
                                       0, & \mbox{otherwise}
                                     \end{array}\right. &&&
\end{align*}}}
\tcp{$\ell_2$ regularization}
Identify the index set $I$, which consists of the elements of $\boldsymbol{\theta}$ corresponding to the rule consequent coefficients, excluding the bias terms\;
\For{each index $i\in I$}{
$\mathbf{g}_k(i)=\mathbf{g}_k(i)+\lambda\cdot \boldsymbol{\theta}_{k-1}(i)$\;}
\tcp{AdaBound}
$\beta_1=0.9$; \quad $\mathbf{m}_k=\beta_1\bm{m}_{k-1}+(1-\beta_1)\bm{g}_k$; \quad $\displaystyle\hat{\bm{m}}_k=\frac{\bm{m}_k}{1-\beta_1^k}$\;
$\beta_2=0.999$; \quad $\bm{v}_k=\beta_2\bm{v}_{k-1}+(1-\beta_2)\bm{g}_k^2$;\quad
$\displaystyle\hat{\bm{v}}_k=\frac{\bm{v}_k}{1-\beta_2^k}$\;
$\displaystyle\hat{\bm{\alpha}}=\max\left[0.01-\frac{0.01}{(1-\beta_2)k+1},
\min\left(0.01+\frac{0.01}{(1-\beta_2)k},
\frac{\alpha}{\sqrt{\hat{\bm{v}}_t}+10^{-8}}\right)\right]$\;
$\bm{\theta}_k=\bm{\theta}_{k-1}-\hat{\bm{\alpha}}\odot\hat{\bm{m}}_k$\;
}
\textbf{Return} $\bm{\theta}_K$
\caption{The MBGD-RDA algorithm for Gaussian TSK fuzzy system optimization. $\odot$ is element-wise product.} \label{alg:MBGD-RDA}
\end{algorithm*}

\subsection{MBGD-RDA Using Trapezoidal MFs}

\begin{figure}[htpb]\centering
\includegraphics[width=.6\linewidth,clip]{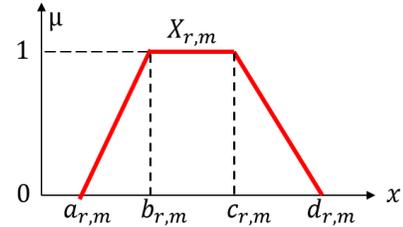}
\caption{A trapezoidal MF $X_{r,m}$, determined by $a_{r,m}$, $b_{r,m}$, $c_{r,m}$ and $d_{r,m}$, where $a_{r,m}<b_{r,m}\le c_{r,m}<d_{r,m}$.} \label{fig:Trap}
\end{figure}

The membership grade of $x_m$ on a trapezoidal MF $X_{r,m}$, shown in Fig.~\ref{fig:Trap}, is:
\begin{align}
\mu_{X_{r,m}}(x_{m})=\left\{\begin{array}{ll}
                              \frac{x_m-a_{r,m}}{b_{r,m}-a_{r,m}}, & x_m\in(a_{r,m},b_{r,m}) \\
                              1, & x\in[b_{r,m},c_{r,m}] \\
                              \frac{d_{r,m}-x_m}{d_{r,m}-c_{r,m}}, & x_m\in(c_{r,m},d_{r,m})\\
                              0, & \mathrm{otherwise}
                            \end{array}
\right.. \label{eq:muT}
\end{align}

When trapezoidal MFs are used, the gradients of the loss function (\ref{eq:loss}) are given in (\ref{eq:dEodCT})-(\ref{eq:dEdBT}).

\begin{figure*}[htpb]
\begin{align}
\frac{\partial L}{\partial a_{r,m}}&=\frac{1}{2}\sum_{x_n\in(a_{r,m},b_{r,m})}\frac{\partial L}{\partial y(\mathbf{x}_n)}
\frac{\partial y(\mathbf{x}_n)}{\partial f_r(\mathbf{x}_n)}
\frac{\partial f_r(\mathbf{x}_n)}{\partial \mu_{X_{r,m}}(x_{n,m})}
\frac{\partial \mu_{X_{r,m}}(x_{n,m})}{\partial a_{r,m}}\nonumber\\
&=\sum_{x_n\in(a_{r,m},b_{r,m})}\left[\left(y(\mathbf{x}_n)-y_n\right)
\frac{y_r(\mathbf{x}_n)\sum_{i=1}^Rf_i(\mathbf{x}_n)
-\sum_{i=1}^Rf_i(\mathbf{x}_n)y_i(\mathbf{x}_n)}
{\left[\sum_{i=1}^Rf_i(\mathbf{x}_n)\right]^2}\frac{f_r(\mathbf{x}_n)}{\mu_{X_{r,m}}(x_{n,m})}
\frac{x_{n,m}-b_{r,m}}{(b_{r,m}-a_{r,m})^2}\right] \label{eq:dEodCT}\\
\frac{\partial L}{\partial b_{r,m}}&=\frac{1}{2}\sum_{x_n\in(a_{r,m},b_{r,m})}\frac{\partial L}{\partial y(\mathbf{x}_n)}
\frac{\partial y(\mathbf{x}_n)}{\partial f_r(\mathbf{x}_n)}
\frac{\partial f_r(\mathbf{x}_n)}{\partial \mu_{X_{r,m}}(x_{n,m})}
\frac{\partial \mu_{X_{r,m}}(x_{n,m})}{\partial b_{r,m}}\nonumber\\
&=\sum_{x_n\in(a_{r,m},b_{r,m})}\left[\left(y(\mathbf{x}_n)-y_n\right)
\frac{y_r(\mathbf{x}_n)\sum_{i=1}^Rf_i(\mathbf{x}_n)
-\sum_{i=1}^Rf_i(\mathbf{x}_n)y_i(\mathbf{x}_n)}
{\left[\sum_{i=1}^Rf_i(\mathbf{x}_n)\right]^2}f_r(\mathbf{x}_n)
\frac{-1}{b_{r,m}-a_{r,m}}\right] \\
\frac{\partial L}{\partial c_{r,m}}&=\frac{1}{2}\sum_{x_n\in(c_{r,m},d_{r,m})}\frac{\partial L}{\partial y(\mathbf{x}_n)}
\frac{\partial y(\mathbf{x}_n)}{\partial f_r(\mathbf{x}_n)}
\frac{\partial f_r(\mathbf{x}_n)}{\partial \mu_{X_{r,m}}(x_{n,m})}
\frac{\partial \mu_{X_{r,m}}(x_{n,m})}{\partial c_{r,m}}\nonumber\\
&=\sum_{x_n\in(c_{r,m},d_{r,m})}\left[\left(y(\mathbf{x}_n)-y_n\right)
\frac{y_r(\mathbf{x}_n)\sum_{i=1}^Rf_i(\mathbf{x}_n)
-\sum_{i=1}^Rf_i(\mathbf{x}_n)y_i(\mathbf{x}_n)}
{\left[\sum_{i=1}^Rf_i(\mathbf{x}_n)\right]^2}f_r(\mathbf{x}_n)
\frac{1}{d_{r,m}-c_{r,m}}\right] \\
\frac{\partial L}{\partial d_{r,m}}&=\frac{1}{2}\sum_{x_n\in(c_{r,m},d_{r,m})}\frac{\partial L}{\partial y(\mathbf{x}_n)}
\frac{\partial y(\mathbf{x}_n)}{\partial f_r(\mathbf{x}_n)}
\frac{\partial f_r(\mathbf{x}_n)}{\partial \mu_{X_{r,m}}(x_{n,m})}
\frac{\partial \mu_{X_{r,m}}(x_{n,m})}{\partial d_{r,m}}\nonumber\\
&=\sum_{x_n\in(c_{r,m},d_{r,m})}\left[\left(y(\mathbf{x}_n)-y_n\right)
\frac{y_r(\mathbf{x}_n)\sum_{i=1}^Rf_i(\mathbf{x}_n)
-\sum_{i=1}^Rf_i(\mathbf{x}_n)y_i(\mathbf{x}_n)}
{\left[\sum_{i=1}^Rf_i(\mathbf{x}_n)\right]^2}\frac{f_r(\mathbf{x}_n)}{\mu_{X_{r,m}}(x_{n,m})}
\frac{x_{n,m}-c_{r,m}}{(d_{r,m}-c_{r,m})^2}\right]\\
\frac{\partial L}{\partial w_{r,m}}&=\frac{1}{2}\sum_{n=1}^{N_{bs}}\frac{\partial L}{\partial y(\mathbf{x}_n)}
\frac{\partial y(\mathbf{x}_n)}{\partial y_r(\mathbf{x}_n)}
\frac{\partial y_r(\mathbf{x}_n)}{\partial w_{r,m}}+\frac{\lambda}{2} \frac{\partial L}{\partial w_{r,m}}=\sum_{n=1}^{N_{bs}}\left[\left(y(\mathbf{x}_n)-y_n\right) \frac{f_r(\mathbf{x}_n)}{\sum_{i=1}^Rf_i(\mathbf{x}_n)}\cdot x_{n,m}\right] +\lambda I(m)w_{r,m}\label{eq:dEdBT}
\end{align}
\end{figure*}

Algorithm~\ref{alg:MBGD-RDA} can still be used to efficiently train a trapezoidal TSK fuzzy system, after making the following three changes:
\begin{enumerate}
\item $k$-means clustering ($k=R$) instead of fuzzy $c$-means clustering should be used in rule initialization. Let $\bar{\bm{c}}_r=[\bar{c}_{r,1},...,\bar{c}_{r,M}]$ be the center of the $r$-th cluster, $\sigma_{r,m}$ the standard deviation of the $m$-th feature in that cluster, and $\bar{y}_r$ the mean of $y_n$ in that cluster. Then, $w_{r,0}=\bar{y}_r$, $w_{r,m}=0$, $a_{r,m}=\bar{c}_{r,m}-10\sigma_{r,m}$, $b_{r,m}=\bar{c}_{r,m}-0.5\sigma_{r,m}$, $c_{r,m}=\bar{c}_{r,m}+0.5\sigma_{r,m}$, $d_{r,m}=\bar{c}_{r,m}+10\sigma_{r,m}$,  $m=1,...,M$. We deliberately make the initial trapezoidal MFs have long legs for two reasons: 1) make sure every point in each input domain is covered by at least one MF, to avoid gap discontinuities \cite{drwuCont2011}; 2) make sure there are enough samples to activate each MF, so that its parameters can be adequately updated, at least at the beginning of the training.
\item Equations (\ref{eq:dEodCT})-(\ref{eq:dEdBT}) should be used in computing the gradients $\bm{g}_k(i)$.
\item In each epoch, after updating, the relationship $a_{r,m}<b_{r,m}\le c_{r,m}<d_{r,m}$ may be violated, so we need to sort them to make sure $a_{r,m}<b_{r,m}\le c_{r,m}<d_{r,m}$ for each $r$ and $m$.
\end{enumerate}

\section{Rule Pruning} \label{sect:pruning}

We propose a very simple yet effective approach for pruning the rules, regardless of the shape of the MFs. The basic idea is to identify rules which are similar enough, combine them, and then re-tune all remaining rules.

Starting from $R_0$ rules, we first compute the $N$ normalized firing levels of each rule on the $N$ training samples, $\bar{\bm{f}}_r=[\bar{f}_r(\bm{x}_1);\ldots;\bar{f}_r(\bm{x}_N)]$, then remove rules whose $\bar{f}_r=\sum_{n=1}^N\bar{f}_r(\bm{x}_n)$ is smaller than $\gamma\mathrm{median}(\{\bar{f}_r\}_{r=1}^R)$, where $\gamma$ is a user-supplied parameter. Let $R$ be the number of remaining rules. Then, we compute the Jaccard similarity measure between the $i$-th and $j$-th rules as:
\begin{align}
s(i,j)=\frac{\min(\bar{\bm{f}}_i,\bar{\bm{f}}_j)}{\max(\bar{\bm{f}}_i,\bar{\bm{f}}_j)}, \quad i,j=1,...,R
\end{align}
We next form a similarity matrix $S\in\mathbb{R}^{R\times R}$, whose $(i,j)$-th element is $s(i,j)$ when $i\neq j$, and 0 when $i=j$. We then identify the two rules $i$ and $j$ with the maximum similarity. If $s(i,j)$ is larger than a threshold $\theta$, then we replace the $i$-th rule by a weighted average of the two, remove the $j$-th rule, replace the $i$-th row (column) of $S$ by the average of the $i$-th and $j$-th rows (columns), remove the $j$-th row and column of $S$, and iterate until no two rules have similarity larger than $\theta$. We next use Algorithm~\ref{alg:MBGD-RDA} to refine the rulebase.

We repeat the above process until the maximum number of rule pruning iterations is reached.

The pseudo-code of our rule pruning algorithm for TSK fuzzy regression models is shown in Algorithm~\ref{alg:prune}. It can be used for both Gaussian and trapezoidal MFs.

\begin{algorithm}[htbp] 
\KwIn{$N$ labeled training samples $\{\mathbf{x}_n,y_n\}_{n=1}^N$, where $\mathbf{x}_n=(x_{n,1},...,x_{n,M})^T\in \mathbb{R}^{M\times1}$; $R_0$, the initial number of rules; $K_0$, the maximum number of training epochs; $\gamma$, the threshold for removing rules with small normalized firing levels; $\theta$, the similarity threshold for combining two rules;
$T$, the number of rule pruning iterations\;
}
\KwOut{A TSK fuzzy regression model with $R\le R_0$ rules.}
$K=\mathrm{round}([0.6K_0, 0.4K_0/(T-1)\cdot \bm{1}_{T-1}])$, where $\bm{1}_{T-1}\in\mathbb{R}^{1\times (T-1)}$ is an all-one vector\;
$R=R_0$, $w_r=1$ ($r=1,...,R$)\;
Train a TSK fuzzy regression model with $R$ rules using Algorithm~\ref{alg:MBGD-RDA} for $K(1)$ epochs\;
\For{$t=2,...,T$}
{Compute the normalized firing levels $\bar{f}_r(\bm{x}_n)$, $r=1,...,R$, $n=1,...,N$\;
Compute $\bar{f}_r=\sum_{n=1}^N\bar{f}_r(\bm{x}_n)$, $r=1,...,R$\;
Remove rules with $\bar{f}<\gamma\cdot\mathrm{median}(\{\bar{f}_r\}_{r=1}^R)$\;
Denote the remaining number of rules as $R$\;
Compute the Jaccard similarity matrix $S\in\mathbf{R}^{R\times R}$ from the $R$ rules\;
\While{the maximum of $S$ is larger than $\theta$}
{Identify $(i,j)$, the location of the maximum of $S$ in its upper-triangular part\;
Parameters of Rule~$i = [w_i ($Parameters of Rule~$i) + w_j($Parameters of Rule~$j)]/(w_i+w_j)$\;
Remove the $j$-th rule from the rulebase\;
$w_i=w_i+1$\;
Remove the $j$-th element from $\bm{w}$\;
Replace the $i$-th row of $S$ by the average of the $i$-th and $j$-th rows, and the $i$-th column by the average of the $i$-th and $j$-th columns\;
Delete the $j$-th row and $j$-th column of $S$\;
$R=R-1$\;}
Refine the remaining $R$ rules using Algorithm~\ref{alg:MBGD-RDA} for $K(t)$ epochs\;
}
\textbf{Return} The final TSK fuzzy regression model with $R$ rules.
\caption{The rule pruning algorithm for TSK fuzzy regression models.} \label{alg:prune}
\end{algorithm}

\section{Experiments} \label{sect:experiments}

This section presents experimental results to demonstrate the effectiveness of the proposed MBGD-RDA variants and the rule pruning algorithm.

\subsection{Datasets}

Ten regression datasets from the CMU StatLib Datasets Archive and the UCI Machine Learning Repository, summarized in Table~\ref{tab:Datasets} and used in \cite{drwuGD2020}, were used again in our experiments.  Same as \cite{drwuGD2020}, each numerical feature was $z$-normalized to have zero mean and unit variance, and the output mean was subtracted.

\begin{table}[htbp] \centering  \setlength{\tabcolsep}{3mm}
\caption{Summary of the 10 regression datasets.}   \label{tab:Datasets}
\begin{tabular}{c|ccc}   \toprule
Dataset    &Source &$N$, no. of samples &$M$, no. of features  \\ \midrule
PM10           &StatLib &500            &7    \\
NO2            &StatLib &500            &7    \\
Housing        &UCI     &506            &13   \\
Concrete       &UCI     &1,030           &8   \\
Airfoil        &UCI     &1,503           &5   \\
Wine-Red       &UCI     &1,599           &11  \\
Abalone        &UCI     &4,177           &8   \\
Wine-White     &UCI     &4,898           &11  \\
PowerPlant     &UCI     &9,568           &4   \\
Protein        &UCI     &45,730          &9   \\ \bottomrule
\end{tabular}
\end{table}

For each dataset, we randomly selected 70\% samples for training, and the remaining 30\% for test. The root mean squared error (RMSE) on the test samples was computed as the performance measure. Each algorithm was repeated 30 times on each dataset, and the average test results are reported.

\subsection{Algorithms}

We compared the performances of the following six algorithms:
\begin{enumerate}
\item The original MBGD-RDA algorithm proposed in \cite{drwuGD2020}. When $M>5$, PCA was used to reduce the dimensionality to 5; otherwise, the original features were used. Two MFs were used for each input. Hence, the total number of rules was $R_0=2^{\min(5,M)}$. This algorithm is denoted as \texttt{MBGD-RDA$^G_{R_0}$} when Gaussian MFs were used, and \texttt{MBGD-RDA$^T_{R_0}$} when trapezoidal MFs were used.

\item Algorithm~\ref{alg:MBGD-RDA} proposed in Section~II, starting with $R_0$ rules and using all original features. It is denoted as \texttt{vMBGD-RDA$^G_{R_0}$} when Gaussian MFs were used, and \texttt{vMBGD-RDA$^T_{R_0}$} when trapezoidal MFs were used.

\item The rule pruning Algorithm~\ref{alg:prune} proposed in Section~III, starting with $R_0$ rules and also using all original features. Two rounds of pruning ($\gamma=0.5$, $\theta=0.5$, $T=3$) were performed, and the remaining number of rules was denoted as $R$. It is denoted as \texttt{vMBGD-RDA$^G_{R_0\rightarrow R}$} when Gaussian MFs were used, and \texttt{vMBGD-RDA$^T_{R_0\rightarrow R}$} when trapezoidal MFs were used.

\item Algorithm~\ref{alg:MBGD-RDA} proposed in Section~II, starting with $R$ (the number of remaining rules after pruning) rules and using all original features. It is denoted as \texttt{vMBGD-RDA$^G_R$} when Gaussian MFs were used, and \texttt{vMBGD-RDA$^T_R$} when trapezoidal MFs were used.

\item The `\emph{anfis}' function in the Matlab 2019b Fuzzy Logic Toolbox, with $R_0$ rules. When fuzzy $c$-means clustering is used in the initialization, only Gaussian MFs can be used. It is denoted as \texttt{ANFIS-GD$^G_{R_0}$} when gradient descent was used as the optimizer, and \texttt{ANFIS-GD-LSE$^G_{R_0}$} when gradient descent plus least squares estimation was used as the optimizer.

\item \texttt{ANFIS-GD$^G_R$} and \texttt{ANFIS-GD-LSE$^G_R$}, which were identical to \texttt{ANFIS-GD$^G_{R_0}$} and \texttt{ANFIS-GD-LSE$^G_{R_0}$}, respectively, except that $R$ rules were used.
\end{enumerate}
The parameters in Algorithm~\ref{alg:MBGD-RDA} were $N_{bs}=64$, $K=500$, $\alpha=0.01$, $\lambda=0.05$ and $P=0.5$, the same as those in \cite{drwuGD2020}. The default parameters in \emph{anfis} were used.

\subsection{Experimental Settings} \label{sect:settings}

Experiments were performed to answer the following three questions:
\begin{enumerate}
\item[\emph{Q1.}] How is the performance of Algorithm~\ref{alg:MBGD-RDA} compared with the state-of-the-art TSK fuzzy system training approaches, e.g., MBGD-RDA in \cite{drwuGD2020} and \emph{anfis} in Matlab 2019b, with the same number of rules? This question can be answered by comparing \texttt{vMBGD-RDA$^G_{R_0}$} with \texttt{MBGD-RDA$^G_{R_0}$} and \texttt{ANFIS-G$_{R_0}$} (all have $R_0$ rules), and \texttt{vMBGD-RDA$^G_R$} with \texttt{ANFIS-GD$^G_R$} and \texttt{ANFIS-GD-LSE$^G_R$} (all have $R$ rules).
\item[\emph{Q2.}] Can the rule pruning algorithm effectively reduce the number of rules, without significantly sacrificing the performance of the TSK fuzzy system? This question can be answered by comparing \texttt{vMBGD-RDA$^G_{R_0}$} with \texttt{vMBGD-RDA$^G_{R_0\rightarrow R}$}, and checking if $R<R_0$.
\item[\emph{Q3.}] If we know $R$, the desired number of rules, should we start from $R_0>R$ rules and then gradually prune them to $R$ rules, or train a TSK fuzzy system directly with $R$ rules? This question can be answered by comparing \texttt{vMBGD-RDA$^G_{R_0\rightarrow R}$} with \texttt{vMBGD-RDA$^G_R$}.
\end{enumerate}

\subsection{Experimental Results with Gaussian MFs}

\begin{figure*}[htbp]\centering
\includegraphics[width=\linewidth,clip]{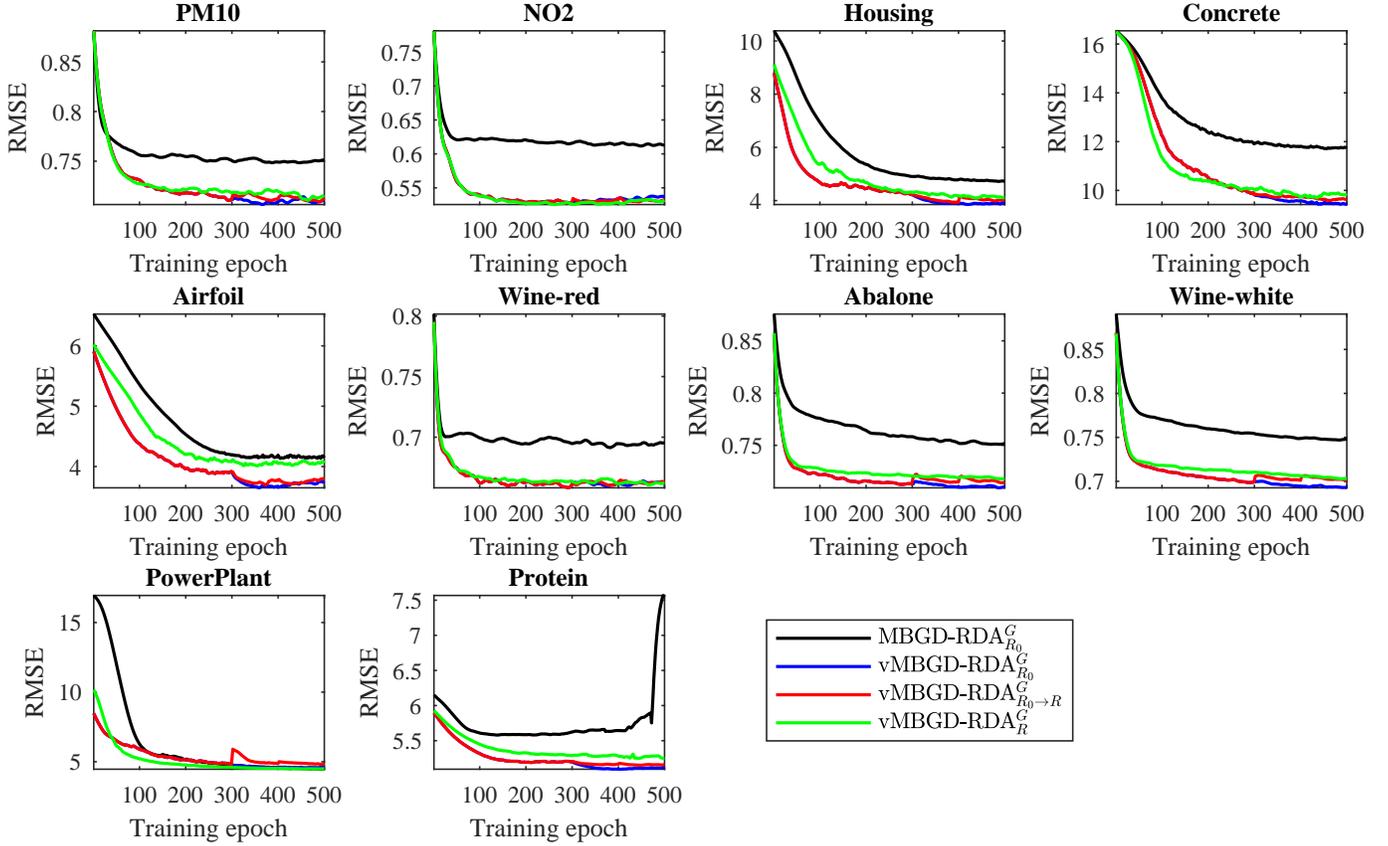}
\caption{Average test RMSEs of the four MBGD based algorithms on the 10 datasets, when Gaussian MFs are used.} \label{fig:mean}
\end{figure*}

The average test RMSEs of the four MBGD based algorithms, when Gaussian MFs are used, are shown in Fig.~\ref{fig:mean}. \texttt{MBGD-RDA$^G_{R_0}$}, the original MBGD-RDA algorithm proposed in \cite{drwuGD2020}, almost always had the worst performance. This is intuitively, because its rules share a lot of common MFs (only two Gaussian MFs were used for each input), and hence the degrees of freedom are small. Assume the TSK fuzzy system has five inputs, and $R_0=32$. Then, \texttt{MBGD-RDA$^G_{R_0}$} has $2\times2\times5=20$ (parameters per MF $\times$ MFs per input $\times$ inputs) antecedent parameters, whereas \texttt{vMBGD-RDA$^G_{R_0}$} has $2\times 5\times 32=320$ (parameters per MF $\times$ MFs per rule $\times$ rules) antecedent parameters. Clearly, the latter is more likely to achieve better performance. This answered the first part of \emph{Q1}: with the same number of rules, our proposed Algorithm~\ref{alg:MBGD-RDA} outperforms the state-of-the-art MBGD-RDA algorithm in \cite{drwuGD2020}.

The remaining number of rules after different rounds of pruning on the 10 datasets are shown in Table~\ref{tab:rulesG}. The two numbers in parenthesis are the number of rules removed by the normalized firing level threshold $\gamma$, and the similarity threshold $\theta$ (see Algorithm~\ref{alg:prune}), respectively. Clearly, $R$ was always smaller than $R_0$, and on many datasets half or more rules were pruned. Both thresholds contributed to the pruning process.

\begin{table}[htbp] \centering  \setlength{\tabcolsep}{3mm}
\caption{Average number of rules (over 30 runs) after different rounds of rule pruning, when Gaussian MFs were used.}   \label{tab:rulesG}
\begin{tabular}{c|ccc}   \toprule
\multirow{2}{*}{Dataset} & \multirow{2}{*}{$R_0$}
&$R$ after first  &$R$ after second   \\
&& round of pruning & round of pruning  \\ \midrule
PM10           &32 &26.5 (4.5, 1.0) & 24.1 (1.9, 0.5)         \\
NO2            &32 &24.6 (5.1, 2.3) & 21.3 (2.7, 0.6)         \\
Housing        &32 & 19.6 (11.1, 1.3) & 15.5 (3.4, 0.8)      \\
Concrete       &32 & 16.2 (15.3, 0.5) & 8.4 (7.6, 0.2)    \\
Airfoil        &32 & 17.3 (13.3, 1.5) & 12.7 (3.3, 1.2)      \\
Wine-Red       &32 & 20.1 (2.5, 9.4) & 16.7 (1.7, 1.7)     \\
Abalone        &32 & 19.1 (3.8, 9.2) & 15.1 (1.6, 2.3)      \\
Wine-White     &32 & 19.6 (3.7, 8.6) & 16.0 (1.3, 2.3)   \\
PowerPlant     &16 & 4.1 (3.1, 8.8) & 3.0 (0.6, 0.5)  \\
Protein        &32 & 21.4 (9.0, 1.6) & 17.0 (3.6, 0.8)      \\ \bottomrule
\end{tabular}
\end{table}

\begin{figure}[htbp]\centering
\includegraphics[width=\linewidth,clip]{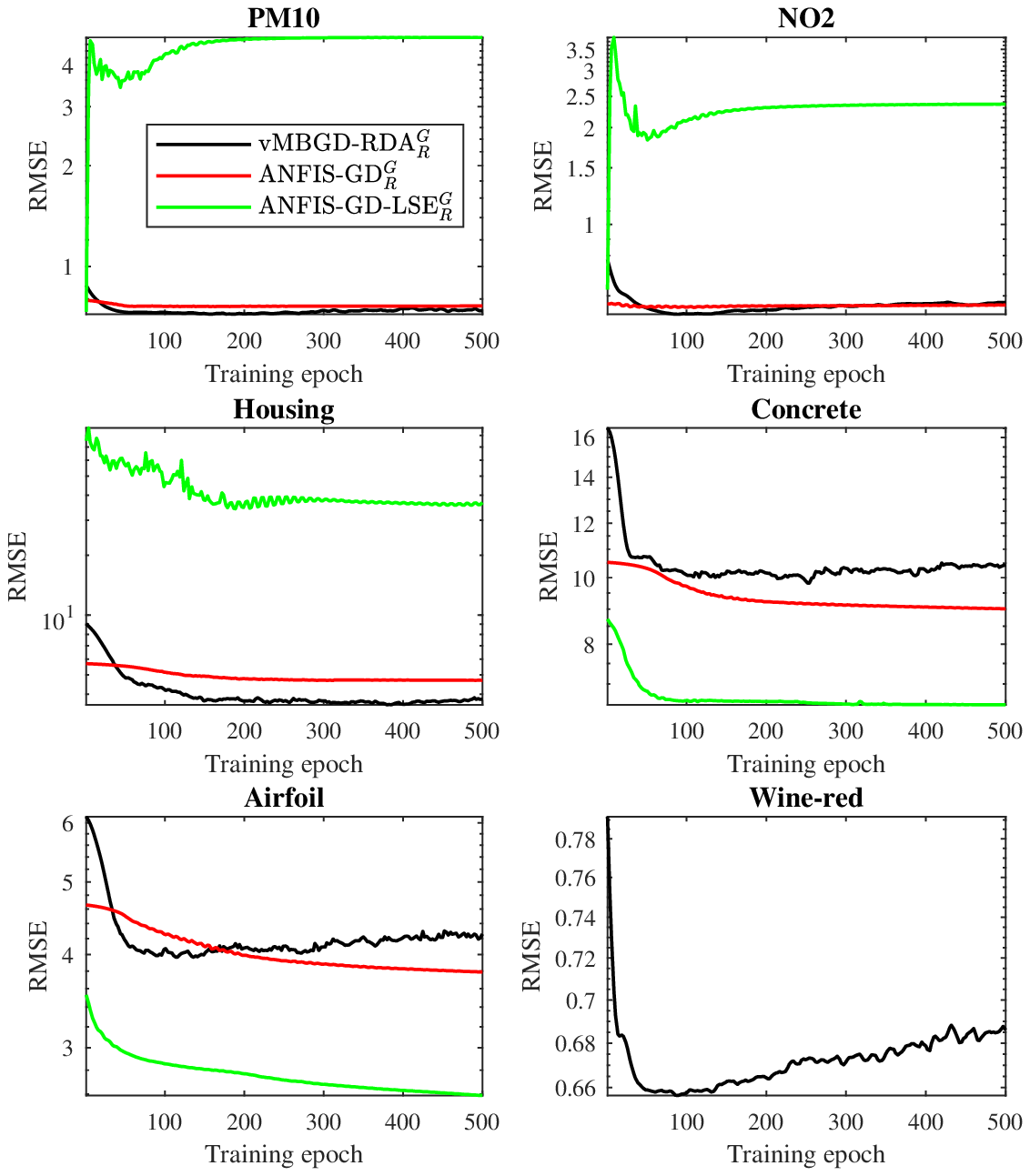}
\caption{Average test RMSEs (over 10 runs) of \texttt{MBGD-RDA$^G_R$}, \texttt{ANFIS-GD$^G_R$} and \texttt{ANFIS-GD-LSE$^G_R$}, when Gaussian MFs were used.} \label{fig:ANFIS}
\end{figure}

Fig.~\ref{fig:ANFIS} shows the performances of \texttt{vMBGD-RDA$^G_R$}, \texttt{ANFIS-GD$^G_R$} and \texttt{ANFIS-GD-LSE$^G_R$}, when $R$ (shown in the last column of Table~\ref{tab:rulesG}) rules were used in all three of them. Because Matlab's \emph{anfis} function always uses the entire training dataset in each training epoch (i.e., the batch size is always $N$), for fair comparison, we also set $N_{bs}=N$ in \texttt{vMBGD-RDA$^G_R$}. This significantly slowed down the training. As a result, we only show the results on the first six smaller datasets in Fig.~\ref{fig:ANFIS}. The performance of \texttt{ANFIS-GD-LSE$^G_R$} was very unstable: sometimes it was much better than the other two, but more likely it was much worse. The performances of \texttt{vMBGD-RDA$^G_R$} and \texttt{ANFIS-GD$^G_R$} were similar on the first five datasets. \texttt{ANFIS-GD$^G_R$} and \texttt{ANFIS-GD-LSE$^G_R$} disappear in the last subfigure, because for unknown reason Matlab's \emph{anfis} function cannot be run at all on the Wine-red dataset. In fact, on the Housing dataset, it also failed three times in 10 runs. These together suggest at least two advantages of our proposed \texttt{vMBGD-RDA$^G_R$} over ANFIS: 1) our algorithm can effectively deal with large datasets, whereas ANFIS cannot; and, 2) our algorithm is more stable than ANFIS. So, the second part of \emph{Q1} is also confirmed: our proposed Algorithm~\ref{alg:MBGD-RDA} outperforms the latest ANFIS algorithm.

\begin{figure*}[htpb]\centering
\includegraphics[width=\linewidth,clip]{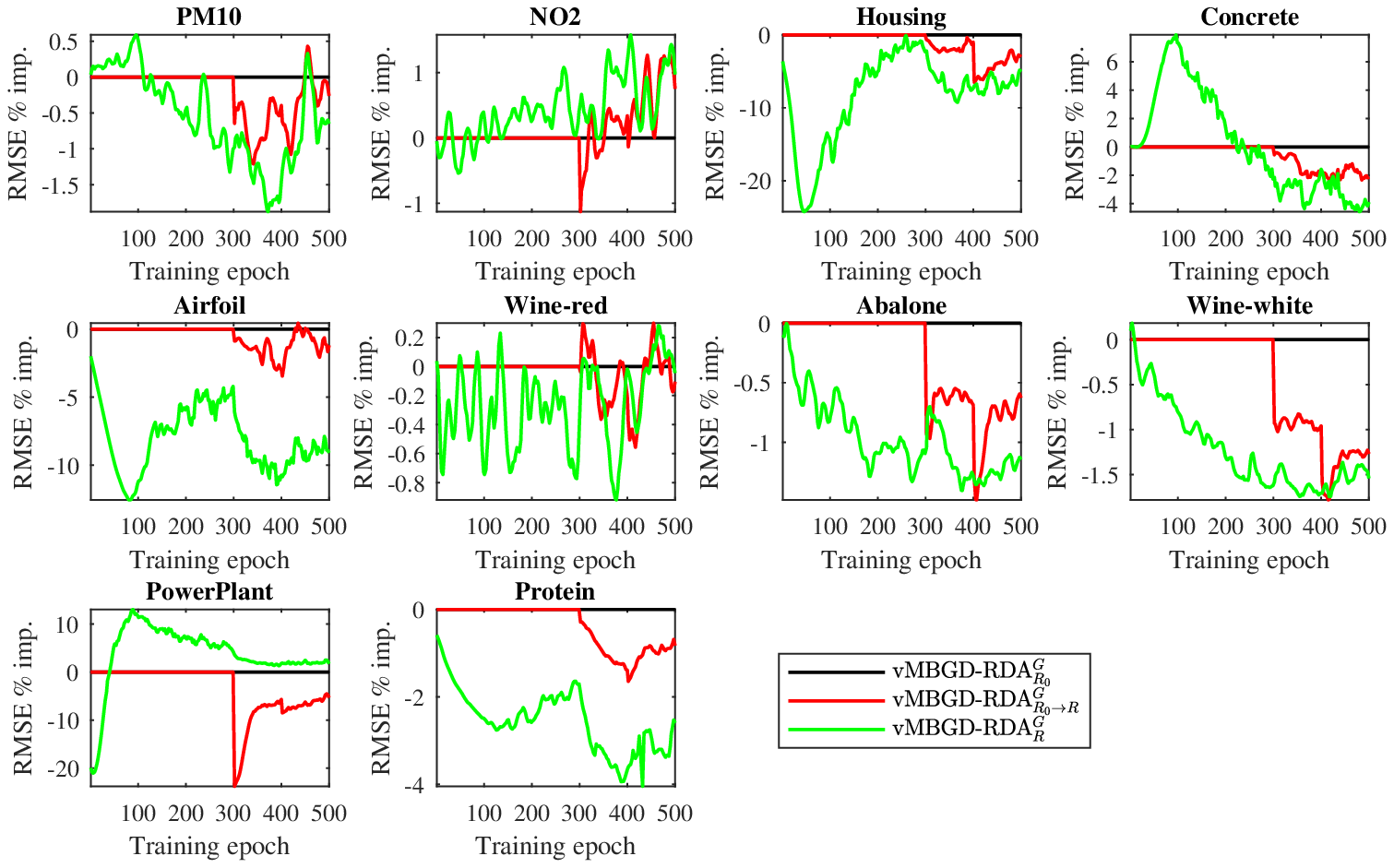}
\caption{Percentage improvements of the average test RMSEs (over 30 runs) of \texttt{vMBGD-RDA$^G_{R_0\rightarrow R}$} and \texttt{vMBGD-RDA$^G_R$} over \texttt{vMBGD-RDA$^G_{R_0}$}.} \label{fig:imp}
\end{figure*}

Fig.~\ref{fig:mean} shows that \texttt{vMBGD-RDA$^G_{R_0}$}, \texttt{vMBGD-RDA$^G_{R_0\rightarrow R}$} and \texttt{vMBGD-RDA$^G_R$} achieved very similar performance at convergence. To better visualize the performance differences among them, as in \cite{drwuGD2020}, we plot in Fig.~\ref{fig:imp} the percentage improvements of \texttt{vMBGD-RDA$^G_{R_0\rightarrow R}$} and \texttt{vMBGD-RDA$^G_R$} over \texttt{vMBGD-RDA$^G_{R_0}$}: in each MBGD training epoch, we treat the test RMSE of \texttt{vMBGD-RDA$^G_{R_0}$} as one, and compute the relative percentage improvements of the test RMSEs of the other two algorithms over it.

Fig.~\ref{fig:imp} shows that on most datasets, the relative performance difference between \texttt{vMBGD-RDA$^G_{R_0}$} and \texttt{vMBGD-RDA$^G_{R_0\rightarrow R}$} was within 1\%, i.e., they had comparable performances. Considering this together with the results in Table~\ref{tab:rulesG}, \emph{Q2} is confirmed: our proposed rule pruning algorithm can effectively reduce the number of rules, without significantly sacrificing the regression performance.

Fig.~\ref{fig:imp} also shows that on most datasets (except PowerPlant), at the end of training (when the number of epochs approaches 500), \texttt{vMBGD-RDA$^G_{R_0\rightarrow R}$} had better performance than, or comparable performance with, \texttt{vMBGD-RDA$^G_R$}. This answers \emph{Q3}: if we know $R$, the desired number of rules, we should start from $R_0>R$ rules and then gradually prune them to $R$ rules, instead of training a TSK fuzzy system directly with $R$ rules. This may suggest that there is a similar Lottery Ticket Hypothesis \cite{Frankle2019,Frankle2020} in the training of TSK fuzzy systems; however, more investigation is needed to confirm that.

In summary, our experiments demonstrated the superiority of the proposed MBGD-RDA variant using Gaussian MFs, and the rule pruning algorithm.

\subsection{Experimental Results with Trapezoidal MFs}

\begin{figure*}[htbp]\centering
\includegraphics[width=\linewidth,clip]{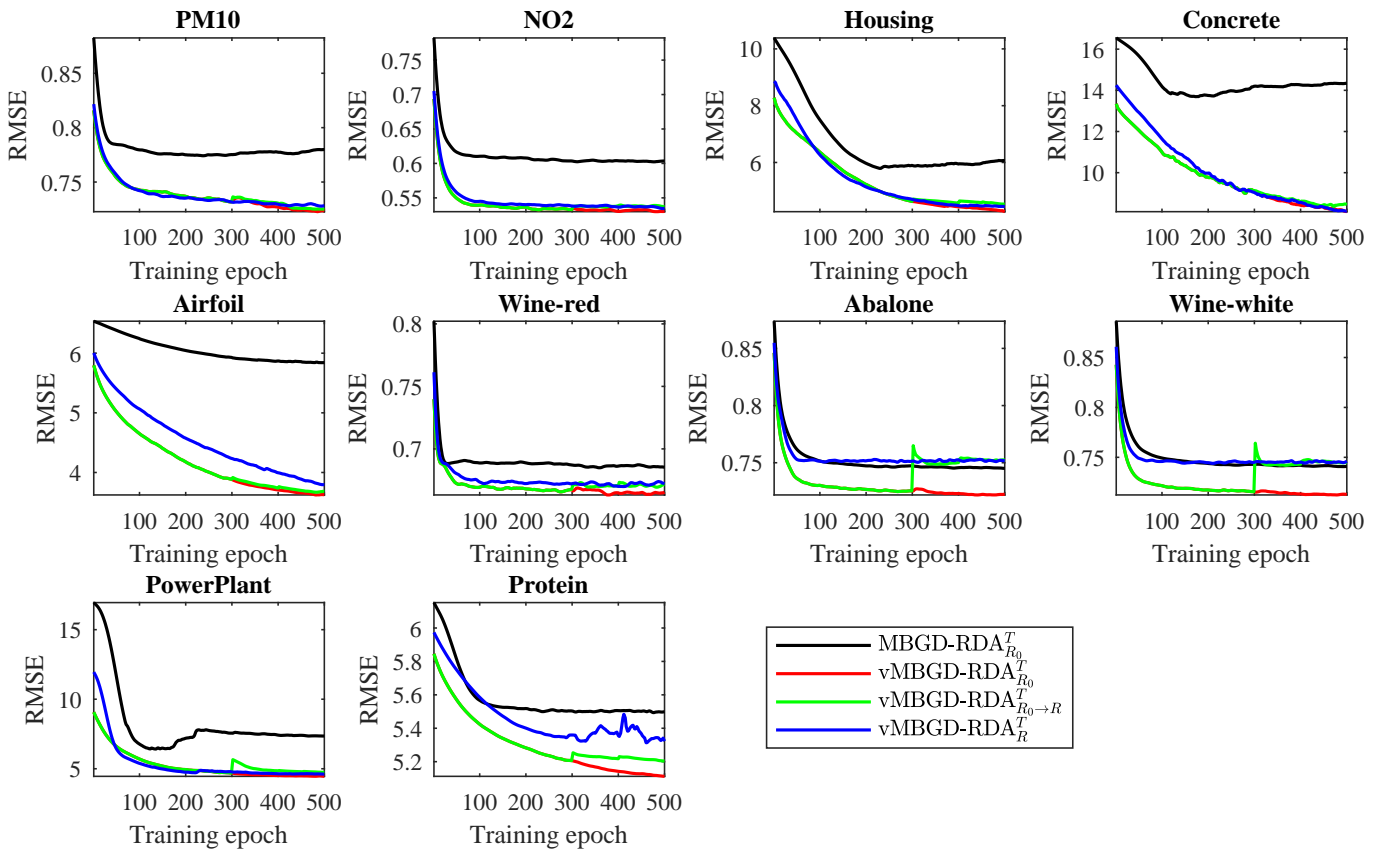}
\caption{Average test RMSEs (over 30 runs) of the four MBGD based algorithms on the 10 datasets, when trapezoidal MFs were used.} \label{fig:meanT}
\end{figure*}

\begin{figure*}[htpb]\centering
\includegraphics[width=\linewidth,clip]{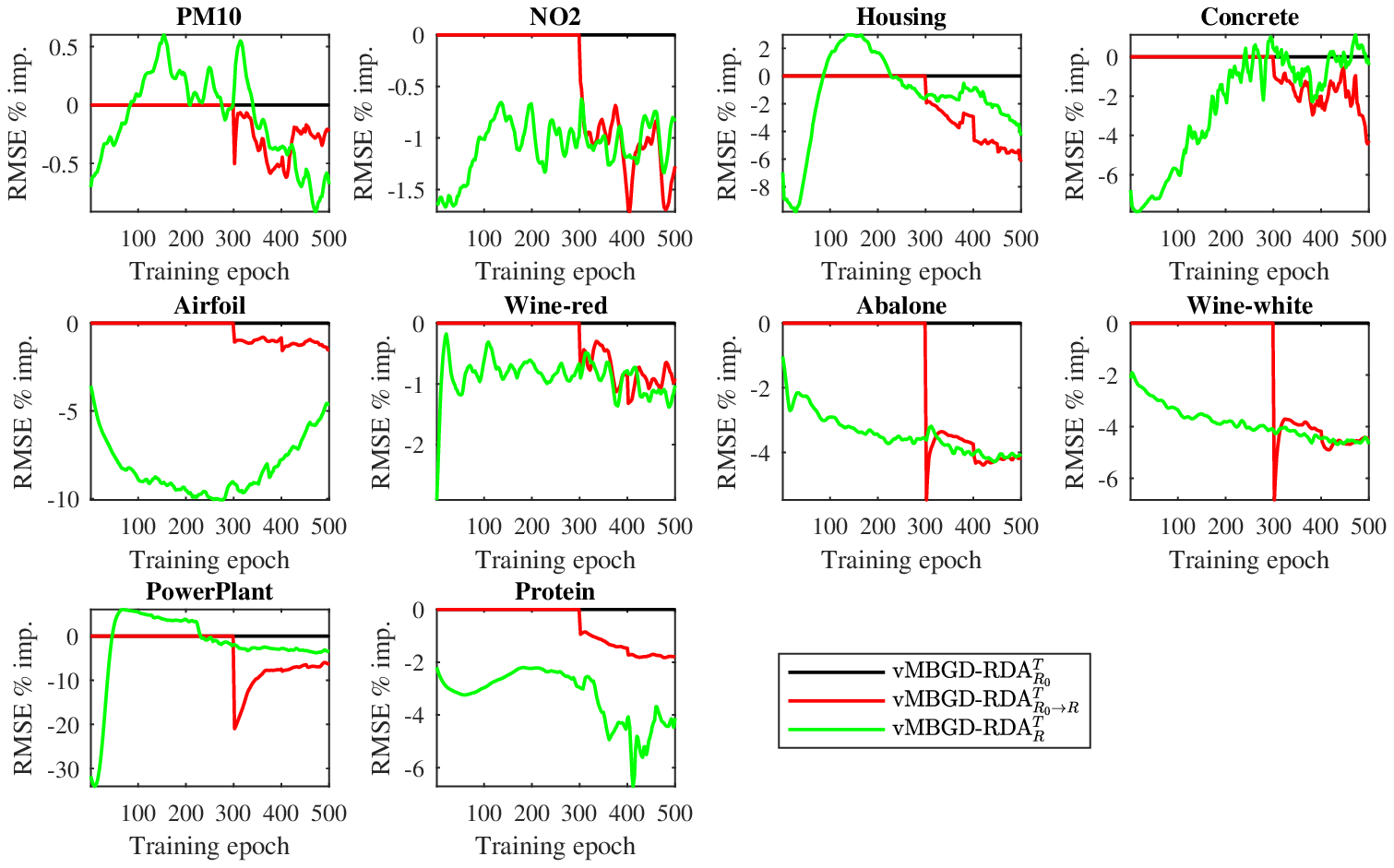}
\caption{Percentage improvements of the average test RMSEs (over 30 runs) of \texttt{vMBGD-RDA$^T_{R_0\rightarrow R}$} and \texttt{vMBGD-RDA$^T_R$} over \texttt{vMBGD-RDA$^T_{R_0}$}.} \label{fig:impT}
\end{figure*}

We also repeated the above experiments for trapezoidal MFs, except the comparisons with ANFIS, because Matlab's \emph{anfis} function does not allow to use fuzzy $c$-means initialization and trapezoidal MFs simultaneously. The results are shown in Figs.~\ref{fig:meanT} and \ref{fig:impT} and Table~\ref{tab:rulesT}. They still provide positive answers to our three questions:
\begin{enumerate}
\item Our proposed Algorithm~\ref{alg:MBGD-RDA} outperformed MBGD-RDA in \cite{drwuGD2020}.
\item Our proposed rule pruning algorithm can effectively reduce the number of rules, without significantly sacrificing the performance of the TSK fuzzy system.
\item Training $R$ rules from pruning a larger rulebase achieved smaller or comparable RMSE than training $R$ rules directly from scratch.
\end{enumerate}

\begin{table}[htbp] \centering  \setlength{\tabcolsep}{3mm}
\caption{Average number of rules (over 30 runs) after different rounds of rule pruning, when trapezoidal MFs were used.}   \label{tab:rulesT}
\begin{tabular}{c|ccc}   \toprule
\multirow{2}{*}{Dataset} & \multirow{2}{*}{$R_0$}
&$R$ after first  &$R$ after second   \\
&& round of pruning & round of pruning  \\ \midrule
PM10           &32 &23.5 (4.1, 4.4) & 22.3 (0.9, 0.3)         \\
NO2            &32 &22.2 (4.1, 5.8) & 20.3 (0.8, 1.1)         \\
Housing        &32 & 20.3 (7.8, 3.9) & 16.6 (2.8, 1.0)      \\
Concrete       &32 & 19.1 (11.3, 1.6) & 15.5 (3.5, 0.1)    \\
Airfoil        &32 & 20.4 (8.5, 3.1) & 17.6 (1.8, 1.0)      \\
Wine-Red       &32 & 11.3 (4.5, 16.2) & 8.4 (0.8, 2.2)     \\
Abalone        &32 & 2.9 (0.8, 28.2) & 2.0 (0.4, 0.6)      \\
Wine-White     &32 & 6.0 (1.8, 24.2) & 5.4 (0.1, 0.6)   \\
PowerPlant     &16 & 4.2 (0.4, 11.4) & 3.8 (0.0, 0.4)   \\
Protein        &32 & 16.2 (5.3 10.5) & 13.7 (1.0, 1.5)      \\ \bottomrule
\end{tabular}
\end{table}

\section{Conclusions} \label{sect:conclusions}

The recently proposed MBGD-RDA algorithm can effectively train TSK fuzzy systems for big data regression problems. However, it does not allow the user to specify the number of rules directly, and only Gaussian MFs can be used. This paper has proposed two variants of MBGD-RDA, in which the user can specify directly the number of rules, and both Gaussian and trapezoidal MFs can be used. These variants outperform the original MBGD-RDA and the classical ANFIS algorithms with the same number of rules. Furthermore, we have proposed an automatic rule pruning algorithm, which can reduce the number of rules without significantly sacrificing the regression performance. Experiments showed that the rules obtained from pruning are generally better than training them from scratch directly, especially when Gaussian MFs are used.


\end{document}